\documentclass[10pt,twocolumn,letterpaper]{article}
\usepackage{cvpr}
\usepackage{times}
\usepackage{epsfig}
\usepackage{graphicx}
\usepackage{amsmath}
\usepackage{amssymb}

\usepackage{color}
\usepackage{bbding}
\usepackage{pifont}
\usepackage{xcolor}
\usepackage{colortbl}
\usepackage{mathtools}
\usepackage{mathrsfs}
\usepackage{multirow}
\usepackage{algorithmic}
\usepackage{subfigure}
\usepackage{array}
\usepackage{tabu}
\usepackage{adjustbox}
\usepackage{tabularx}
\usepackage{booktabs}
\usepackage{diagbox}
\usepackage{array}
\usepackage{epstopdf}
\usepackage{enumitem}
\usepackage{stfloats}
\usepackage{authblk}

\newcommand{\textBC}[2]{\textbf{\textcolor{#1}{#2}}}

\usepackage[pagebackref=true,breaklinks=true,letterpaper=true,colorlinks,bookmarks=false]{hyperref}

\cvprfinalcopy 

\def\cvprPaperID{} 

\ifcvprfinal\fi
\begin{document}

\title{Multi-scale Interactive Network for Salient Object Detection}

\setlength{\affilsep}{0em}

\author{
Youwei Pang$^1$\protect\footnotemark[2]\,, Xiaoqi Zhao$^1$\protect\footnotemark[2]\,, Lihe Zhang$^1$\protect\footnotemark[1]\, and Huchuan Lu$^{1,2}$\\
$^1$Dalian University of Technology, China \\
$^2$Peng Cheng Laboratory \\
{\tt\small \{lartpang, zxq\}@mail.dlut.edu.cn, \{zhanglihe, lhchuan\}@dlut.edu.cn}
}

\maketitle
\thispagestyle{empty}

\renewcommand{\thefootnote}{\fnsymbol{footnote}} 
\footnotetext[2]{These authors contributed equally to this work.} 
\footnotetext[1]{Corresponding author.} 
\renewcommand{\thefootnote}{\arabic{footnote}}

\begin{abstract}
    Deep-learning based salient object detection methods achieve great progress. However, the variable scale and unknown category of salient objects are great challenges all the time. These are closely related to the utilization of multi-level and multi-scale features. In this paper, we propose the aggregate interaction modules to integrate the features from adjacent levels, in which less noise is introduced because of only using small up-/down-sampling rates. To obtain more efficient multi-scale features from the integrated features, the self-interaction modules are embedded in each decoder unit. Besides, the class imbalance issue caused by the scale variation weakens the effect of the binary cross entropy loss and results in the spatial inconsistency of the predictions. Therefore, we exploit the consistency-enhanced loss to highlight the fore-/back-ground difference and preserve the intra-class consistency. Experimental results on five benchmark datasets demonstrate that the proposed method without any post-processing performs favorably against 23 state-of-the-art approaches. The source code will be publicly available at https://github.com/lartpang/MINet.
\end{abstract}

\vspace{-1em}
\section{Introduction}

\begin{figure}[t]
    \centering
    \includegraphics[width=\linewidth]{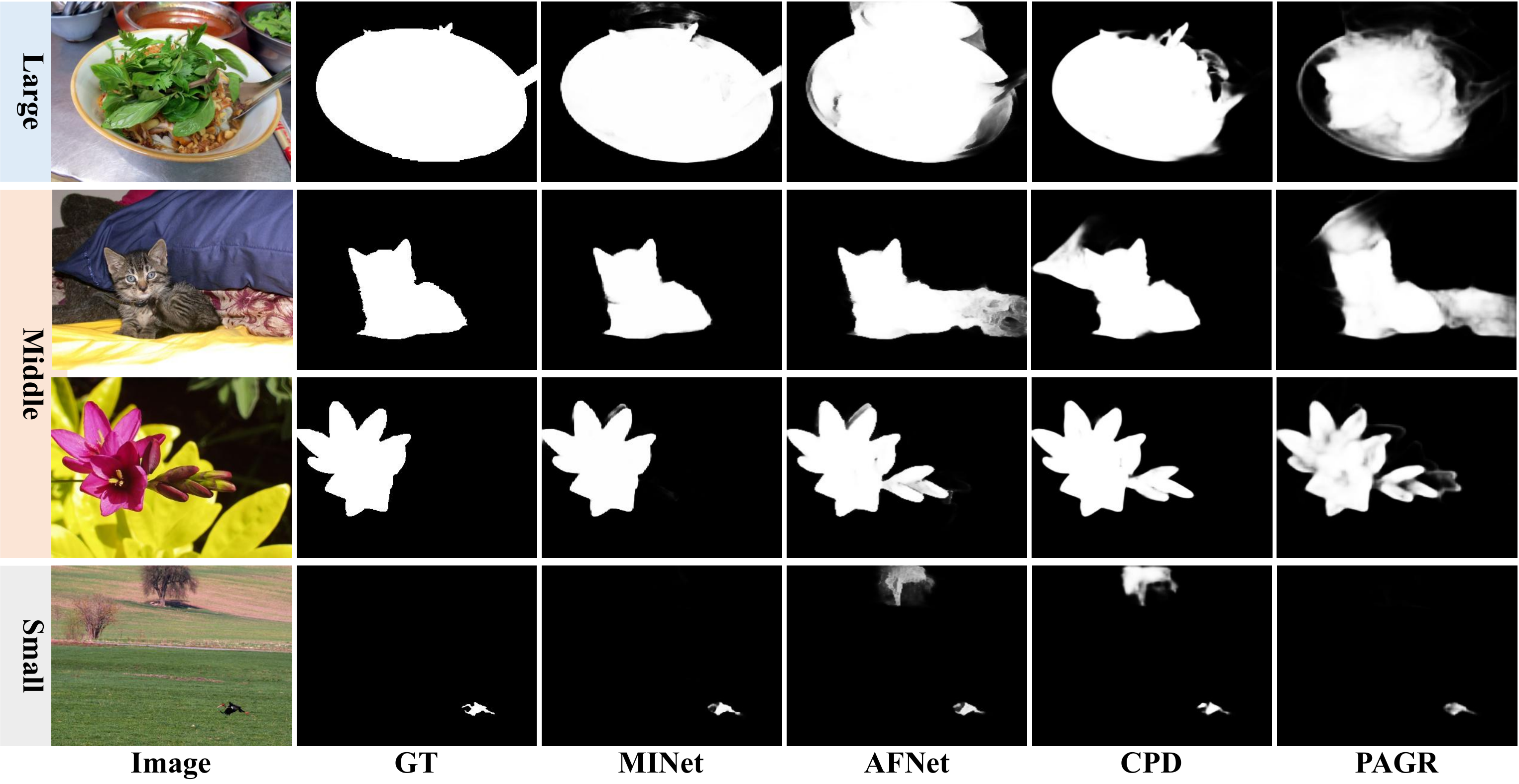}
    \caption{Several visual examples with size-varying objects and their predictions generated by the proposed MINet, AFNet~\cite{AFNetRGB}, CPD~\cite{CPD} and PAGR~\cite{PAGRN} methods.}
    \label{fig:hardexapmles}
    \vspace{-0.5em}
\end{figure}

Salient object detection (SOD) aims at distinguishing the most visually obvious regions. It is growing rapidly with the help of data-driven deep learning methods and has been applied in many computer vision fields, such as
visual tracking~\cite{tracking},
image retrieval~\cite{Retrieval},
non-photorealistic rendering~\cite{nonphotorender},
4D saliency detection~\cite{DeepLightField},
no-reference synthetic image quality assessmen~\cite{NSImageQualityAssessment}
and so on. Although great progress has been made at present, two issues still need to be paid attention to \textit{how to extract more effective information from the data of scale variation} and \textit{how to improve the spatial coherence of predictions in this situation}. Due to various scales of salient regions, the CNN-based methods, which are limited by the absence of necessary detailed information owing to the repeated sub-sampling, have difficulty in consistently and accurately segmenting salient objects of different scales (Fig.~\ref{fig:hardexapmles}). In addition, on account of the inherent localization of convolution operation and the pixel-level characteristics of the cross entropy function, it is difficult to achieve uniform highlighting of objects.

\begin{figure}[t]
    \centering
    \includegraphics[width=\linewidth]{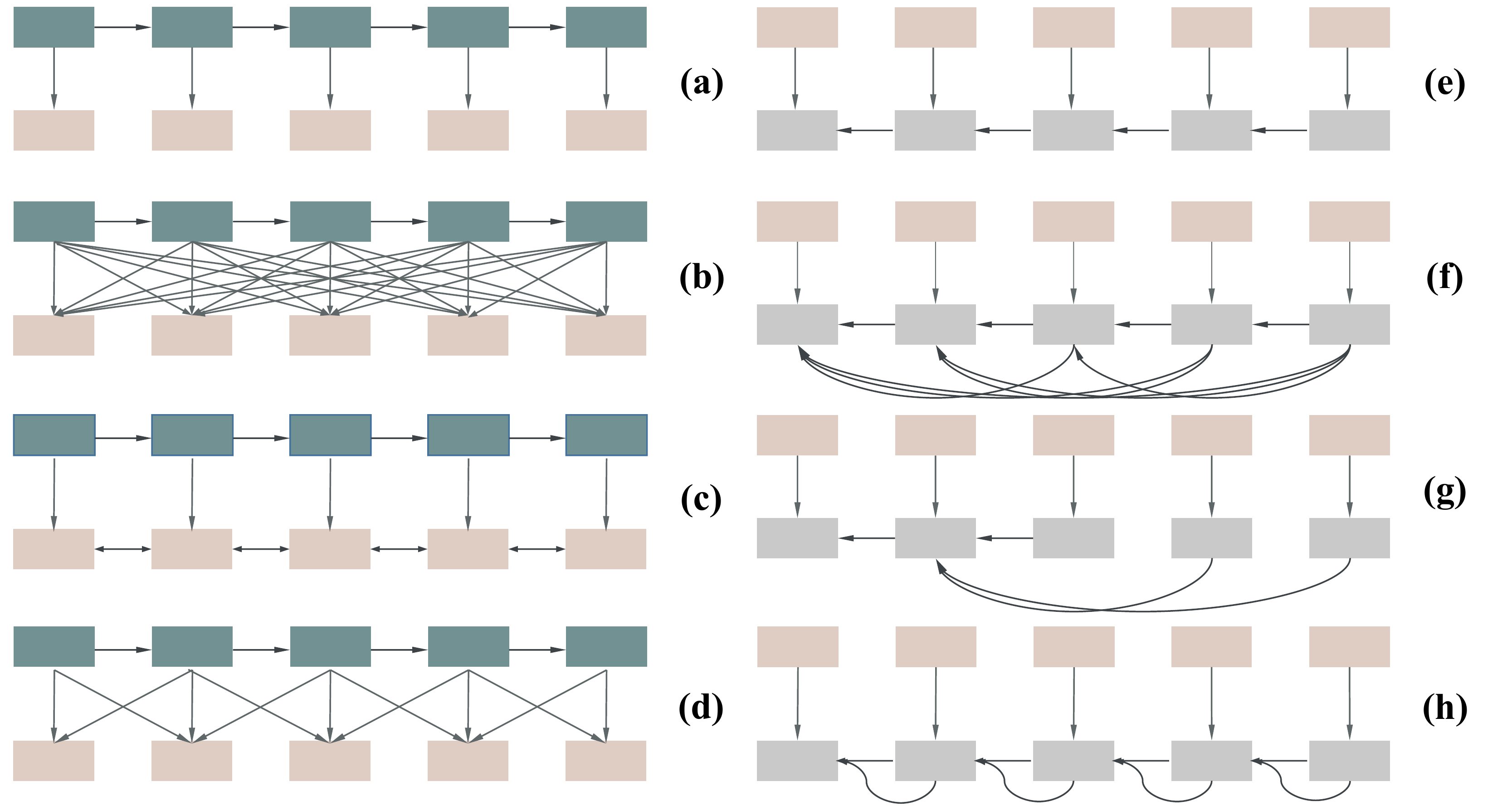}
    \caption{Illustration of different architectures. Green blocks, orange blocks and gray blocks respectively denote the different convolutional blocks in the encoder, the transport layer and the decoder. Left column: the connection patterns between the encoder and the transport layer; Right column: the connection patterns between the transport layer and the decoder. (a, e) FCN~\cite{FCN}; (b) Amulet~\cite{Amulet}; (c) BMPM~\cite{BMPM}; (d) AIMs (Sec.~\ref{sec:aim}); (f) DSS~\cite{DSS}; (g) DGRL~\cite{DGRL}; (h) SIMs (Sec.~\ref{sec:sim}).}
    \label{fig:connectionmodel}
    \vspace{-0.5em}
\end{figure}

For the first problem, the main solution of the existing methods is to layer-by-layer integrate shallower features.
Some methods~\cite{NLDF,PAGRN,RAS,AFNetRGB,MLMSNet,CPD,BASNet,PAGE-Net} connect the features at the corresponding level in the encoder to the decoder via the transport layer (Fig.~\ref{fig:connectionmodel}(a, c, e)).
The single-level features can only characterize the scale-specific information. In the top-down pathway, the representation capability of details in shallow features is weakened due to the continuous accumulation of the deeper features.
To utilize the multi-level features, some approaches~\cite{Amulet, DSS, DGRL} combine the features from multiple layers in a fully-connected manner or a heuristic style (Fig.~\ref{fig:connectionmodel}(b, f, g)).
However, integrating excessive features and lacking a balance between different resolutions easily lead to high computational cost, plenty of noise and fusion difficulties, thereby disturbing the subsequent information recovery in the top-down pathway.
Moreover, the atrous spatial pyramid pooling module (ASPP)~\cite{Deeplab} and the pyramid pooling module (PPM)~\cite{PPM} are used to extract multi-scale context-aware features and enhance the single-layer representation~\cite{R3Net, SRM}.
Nonetheless, the existing methods usually equip these modules behind the encoder, which results in that their networks miss many necessary details due to the limitation of the low resolution of the top-layer features.
For the second problem, some existing models~\cite{CPD, BASNet} mainly employ a specific branch or an additional network to refine the results. Nevertheless, these methods are faced with the problem of computational redundancy and training difficulties, which is not conducive to further applications.

Inspired by the idea of the mutual learning proposed by Zhang et al.~\cite{DML}, we propose an aggregate interaction strategy (AIM) to make better use of multi-level features and avoid the interference in feature fusion caused by large resolution differences (Fig.~\ref{fig:connectionmodel}(d)). We collaboratively learn knowledge guidance to effectively integrate the contextual information from adjacent resolutions.
To further obtain abundant scale-specific information from the extracted features, we design a self-interaction module (SIM) (Fig.~\ref{fig:connectionmodel}(h)). Two interactive branches of different resolutions are trained to learn multi-scale features from a single convolutional block. AIMs and SIMs effectively improve the ability to deal with scale variations in the SOD task.
Unlike the settings in~\cite{DML}, in the two modules, the mutual learning mechanism is incorporated into feature learning. Each branch can more flexibly integrate information from other resolutions through interactive learning. In AIMs and SIMs, the main branch ($\textit{B}^1$ in Fig.~\ref{fig:aim} and $\textit{B}^0$ in Fig.~\ref{fig:sim}) is supplemented by the auxiliary branches and its discriminating power is further enhanced.
In addition, the multi-scale issue also causes a serious imbalance between foreground and background regions in the datasets, hence we embed a consistency-enhanced loss (CEL) into the training stage, which is not sensitive to the scale of objects. At the same time, the CEL can better handle the spatial coherence issue and uniformly highlight salient regions without additional parameters, because its gradient has the characteristics of keeping intra-class consistency and enlarging inter-class differences.

Our contributions are summarized as three folds:

\begin{itemize}[noitemsep, nolistsep]
    \item We propose the MINet to effectively meet scale challenges in the SOD task. The aggregate interaction module can efficiently utilize the features from adjacent layers by the way of mutual learning and the self-interaction module makes the network adaptively extract multi-scale information from data and better deal with scale variation.
    \item We utilize the consistency-enhanced loss as an assistant to push our model to uniformly highlight the entire salient region and better handle the pixel imbalance problem between fore- and back-ground regions caused by various scales of objects, without any post-processing or extra parameters.
    \item We compare the proposed method with 23 state-of-the-art SOD methods on five datasets. It achieves the best performance under different evaluation metrics. Besides, the proposed model has a forward reasoning speed of 35 FPS on GPU.
\end{itemize}

\section{Related Work}

\subsection{Salient Object Detection}

Early methods are mainly based on hand-crafted priors~\cite{globalcontrast,geodesic,RankingSaliency,AbsorbingMarkovChainSOD}. Their generalization and effectiveness are limited. The early deep salient object detection (SOD) methods~\cite{multicontext, HKU-IS} use the multi-layer perception to predict the saliency score for each processing unit of an image. These methods have low computational efficiency and damage the potential feature structure. See~\cite{CMMSODSurvey, WWGSODSurvey} for more details about traditional and early deep methods.

Recently, some methods~\cite{PiCANet, PAGRN} introduce the fully convolutional network (FCN)~\cite{FCN} and achieve promising results. Moreover, Liu et al.~\cite{PiCANet} hierarchically embed global and local context modules into the top-down pathway which constructs informative contextual features for each pixel.
Chen et al.~\cite{RAS} propose reverse attention in the top-down pathway to guide residual saliency learning, which drives the network to discover complement object regions and details.
Nonetheless, the above-mentioned methods only employ individual resolution features in each decoder unit, which is not an effective enough strategy to cope with complex and various scale problems.

\begin{figure*}
    \centering
    \includegraphics[width=\linewidth]{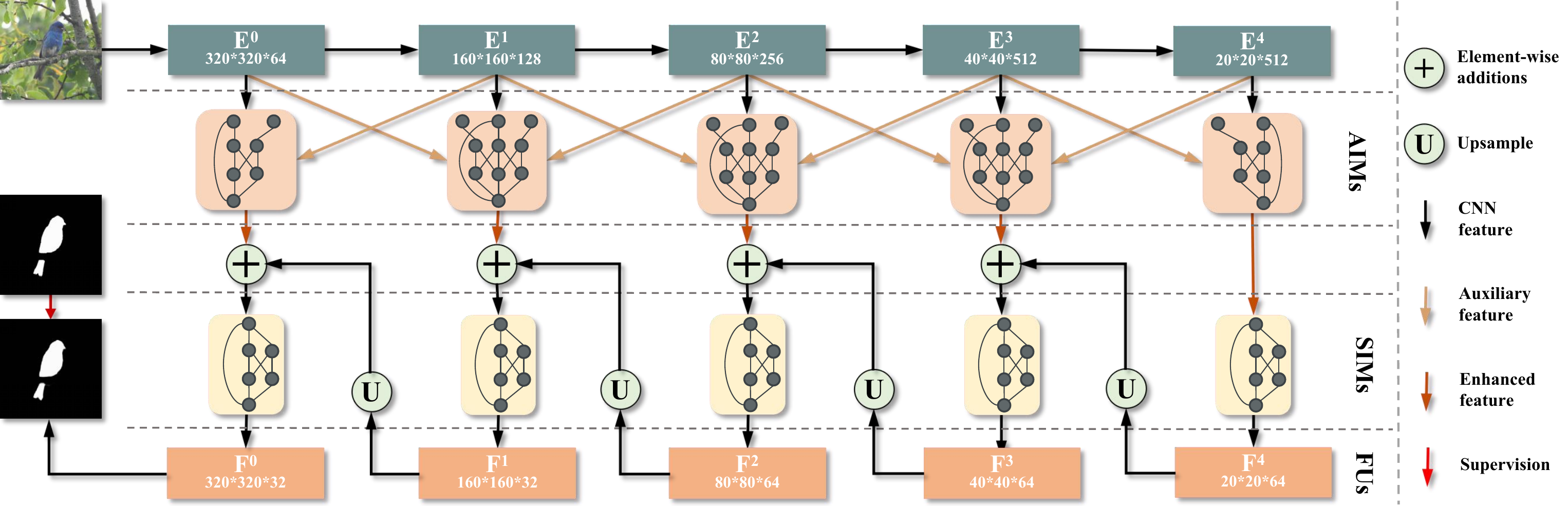}
    \caption{The overall framework of the proposed model. Each colorful box represents a feature processing module. Our model takes a RGB image ($320 \times 320 \times 3$) as input, and exploits VGG-16~\cite{VGG} blocks $\{\mathbf{E}^i\}_{i=0}^4$ to extract multi-level features. The features are integrated by AIMs ($\{\mathbf{AIM}^i\}_{i=0}^4$) and then, the outputted features are gradually combined by using SIMs ($\{\mathbf{SIM}^i\}_{i=0}^4$) and fusion units ($\{\mathbf{F}^i\}_{i=0}^4$) to generate the final prediction $\mathcal{P}$ supervised by the ground truth $\mathcal{G}$.}
    \label{fig:minet}
\end{figure*}

\subsection{Scale Variation}

Scale variation is one of the major challenges in the SOD task. Limited by the localized convolution operation and sub-sampling operation, it is difficult for CNN to handle this problem.
On one hand, the amount of information about objects, which are embedded in the features of different resolutions, changes with the scale of objects.
A straightforward strategy is to roughly integrate all features.
On the other hand, each convolutional layer only has the capability of processing a special scale. Therefore, we need to characterize the multi-scale information from a single layer by building a multi-path feature extraction structure.

\noindent\textbf{Multi-level Information.} Zhang et al.~\cite{Amulet} simply combine all level features into the transport layer. This kind of coarse fusion easily produces information redundancy and noise interference. In~\cite{BMPM}, a gate function is exploited to control the message passing rate to optimize the quality of information exchange between layers. Nevertheless, multiple gating processing leads to severe attenuation of the information from other layers, which limits the learning ability of the network. Different from these methods, we only fuse the features from the adjacent layers by reason of their closer degree of abstraction and concurrently obtain abundant scale information.

\noindent\textbf{Multi-scale Information.} The atrous spatial pyramid pooling (ASPP)~\cite{Deeplab} and the pyramid pooling module (PPM)~\cite{PPM} are two common choices for multi-scale information extraction and are often fixed at the deepest level in the network~\cite{R3Net, SRM}. Since the deeper features contain less information about small-scale objects, which is especially true for the top-layer features, these methods can not effectively deal with large scale variation.
Besides, in~\cite{PAGE-Net}, the pyramid attention module can obtain multi-scale attention maps to enhance features through multiple downsampling and softmax operations on all positions. But such a softmax severely suppresses non-maximum values and is more sensitive to noise. It does not improve the scale issue well.
To avoid misjudging small objects, we propose a multi-scale processing module where two branches interactively learn features. Through data-driven training, the two-path structure can learn rich multi-scale representation.
In addition, the oversized and undersized objects cause the imbalance between foreground and background samples, which weakens the effect of pixel-level supervision. We introduce the consistency-enhanced loss (CEL) as an aid to the cross entropy loss. The CEL is not sensitive to the size of objects. It can overcome the difficulties of supervision and perform very well in the face of large scale variation.

\subsection{Spatial Coherence}

To improve spatial coherence and quality of saliency maps, some non-deep methods often integrate an over-segmentation process that generates regions~\cite{ECSSD}, super-pixels~\cite{DUT-OMRON}, or object proposals~\cite{VideoProposal}. For deep learning based methods, Wu et al.~\cite{CPD} propose a cascaded partial decoder framework with two branches and directly utilize attention maps generated by the attention branch to refine the features from the saliency detection branch. Qin et al.~\cite{BASNet} employ a residual refinement module combined with a hyper loss to further refine the predictions, which significantly reduces the inference speed. In this paper, the CEL pays more attention to the overall effect of the prediction. It helps obtain a more uniform saliency result and is a better trade-off between the effect and the speed.

\section{Proposed Method}\label{sec:proposedmthod}

In this paper, we propose an interactive integration network which fuses multi-level and multi-scale feature information to deal with the prevalent scale variation issue in saliency object detection (SOD) task. The overall network structure is shown in Fig.~\ref{fig:minet}. Encoder blocks, aggregate interaction modules (AIMs), self-interaction modules (SIMs) and fusion units (FUs) are denoted as $\{\mathbf{E}^i\}_{i=0}^4$, $\{\mathbf{AIM}^ i\}_{i=0}^4$, $\{\mathbf{SIM}^i\}_{i=0}^ 4$ and $\{\mathbf{F}^i\}_{i=0}^4$, respectively.

\subsection{Network Overview}\label{sec:networkoverview}

Our model is built on the FCN architecture with the pretrained VGG-16~\cite{VGG} or ResNet-50~\cite{Resnet} as the backbone, both of which only retain the feature extraction network. Specifically, we remove the last max-pooling layer of the VGG-16 to maintain the details of the final convolutional layer. Thus, the input is sub-sampled with a factor of $16$ for the VGG-16 and with a factor of $32$ for the ResNet-50. We use the backbone to extract multi-level features and abstractions, and then each AIM (Fig.~\ref{fig:aim}) uses the features of adjacent layers as the input to efficiently employ the multi-level information and provide more relevant and effective supplementary for the current resolution. Next, in the decoder, every SIM (Fig.~\ref{fig:sim}) is followed by an FU which is a combination of a convolutional layer, a batch normalization layer and a ReLU layer. The SIM can adaptively extract multi-scale information from the specific levels. The information is further integrated by the FU and fed to the shallower layer.
In addition, we introduce the consistency-enhanced loss as an auxiliary loss to supervise the training stage. In this section, we will introduce these modules in detail. To simplify the description, all subsequent model parameters are based on the VGG-16 backbone.

\subsection{Aggregate Interaction Module}\label{sec:aim}

\begin{figure}[t]
    \centering
    \includegraphics[width=1 \linewidth]{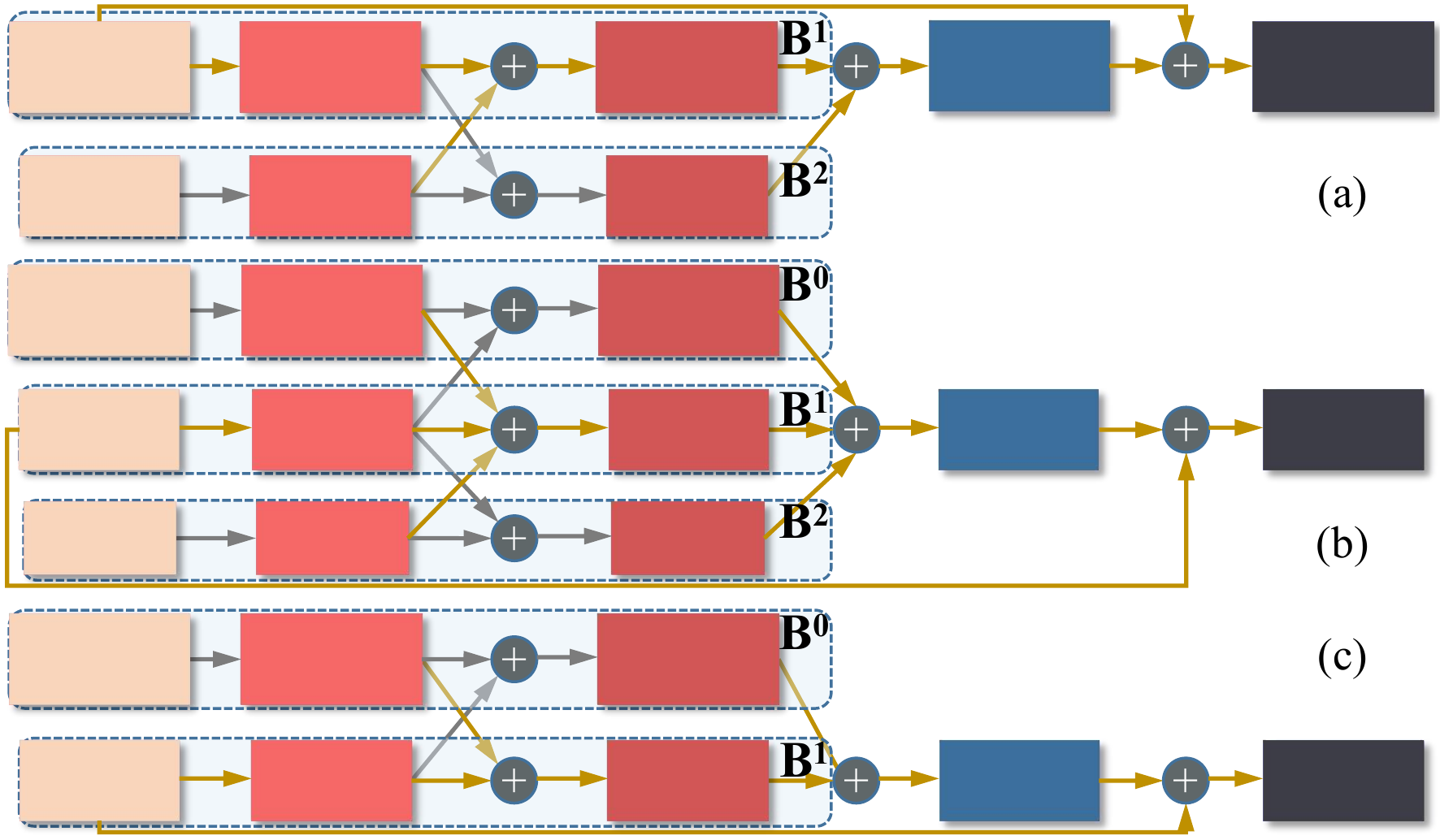}
    \caption{Illustration of aggregate interaction modules (AIMs). $\textbf{B}^{i}$: All operations in the $i^{th}$ branch $\textit{B}^{i}$.}
    \label{fig:aim}
\end{figure}

In the feature extraction network, different levels of convolutional layers correspond to a different degree of feature abstraction. The multi-level integration can enhance the representation ability of different resolution features: 1) In the shallow layers, the detailed information can be further strengthened and the noise can be suppressed; 2) In the middle layers, both semantic and detailed information is taken into account at the same time, and the proportion of different abstraction information in the features can be adaptively adjusted according to the needs of the network itself, thereby achieving more flexible feature utilization; 3) In the top layer, richer semantic information can be mined when considering adjacent resolutions.
In particular, we propose the aggregate interaction module (AIM) (Fig.~\ref{fig:aim}) to aggregate features by a strategy of interactive learning.

The $i^{th}$ AIM is denoted as $\mathbf{AIM}^i$, the input of which consists of features $f_e^{i-1}$, $f_e^i$ and $f_e ^{i+1}$ from the encoder, as shown in Fig.~\ref{fig:aim} (b). After the initial \textbf{transformation} by a combination of a single convolutional layer, a batch normalization layer and a ReLU layer, the channel number of these features is reduced. In the \textbf{interaction} stage, the $\textit{B}^0$ branch and the $\textit{B}^2$ branch are adjusted by the pooling, neighbor interpolation and convolution operations, and then both of them are merged into the $\textit{B}^1$ branch by the element-wise addition. At the same time, the $\textit{B}^1$ branch is also adjusted its resolution and is respectively combined into the $\textit{B}^0$ and $\textit{B}^2$ branches. Finally, the three branches are \textbf{fused} together through the subsequent convolutional layer and the channel number is also reduced. In order to efficiently train the AIMs and increase the weight of $f_e^i$ to ensure that other branches only act as supplements, a residual learning strategy is introduced. The outputted feature is denoted as $f_{AIM}^i \in \mathbb{R}^{N_i \times H_i \times W_i \times C_i}$, where $C_0 = 32$ and $C_{i\neq 0} = 64$. For $\mathbf{AIM}^0$ and $\mathbf{AIM}^4$ , their inputs only contain $f_e^0$, $f_e^1$ and $f_e^3$, $f_e^4$, correspondingly (Fig.~\ref{fig:aim} (a, c)). The entire process is formulated as follows:
\begin{equation}
    \begin{split}
        f_{AIM}^i & = \textbf{I}_{AIM}^i(f_e^i) + \textbf{M}_{AIM}^i(f_{AB}^i), \\
        f_{AB}^i & = \left \{ \begin{matrix}
            \sum_{j=1}^2 \textbf{B}_{AIM}^{i,j}(f_e^{j-1})   & \text{ if } i=0,       \\
            \sum_{j=0}^2 \textbf{B}_{AIM}^{i,j}(f_e^{i+j-1}) & \text{ if } i=1, 2, 3, \\
            \sum_{j=0}^1 \textbf{B}_{AIM}^{i,j}(f_e^{i+j-1}) & \text{ if } i=4,
        \end{matrix} \right.
    \end{split}
    \label{equ:aim}
\end{equation}

\noindent where $\textbf{I}(\cdot)$ and $\textbf{M}(\cdot)$ represent the identity mapping and the branch merging, respectively. $\textbf{B}_{AIM}^{i,j}(\cdot)$ is the overall operation of the $j^{th}$ branch (i.e.\ $\textit{B}^j$) in the $\mathbf{AIM}^i$. Due to space constraints, please refer to Fig.~\ref{fig:aim} for the computational details inside each branch.

\subsection{Self-Interaction Module}\label{sec:sim}

\begin{figure}[t]
    \centering
    \includegraphics[width=1 \linewidth]{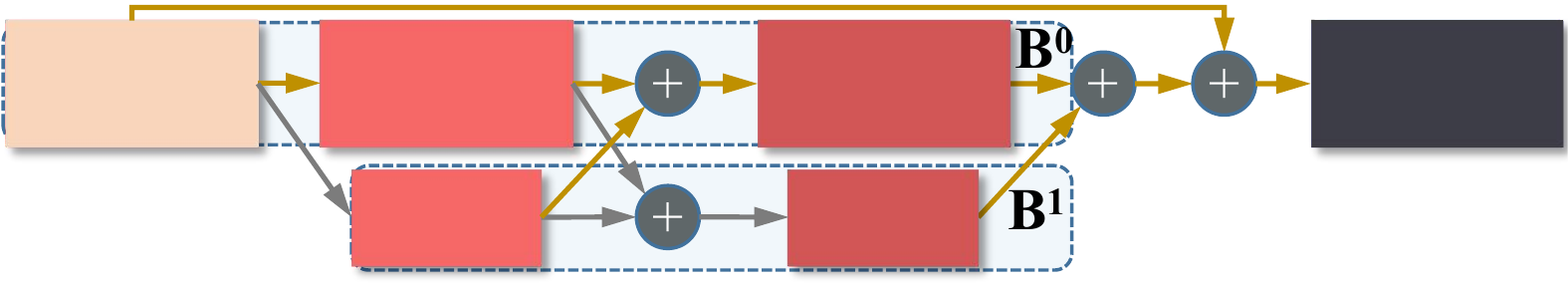}
    \caption{Illustration of self-interaction modules (SIMs). $\textbf{B}^{i}$: All operations in the $i^{th}$ branch $\textit{B}^{i}$.}
    \label{fig:sim}
\end{figure}

The AIMs aim at achieving efficient utilization of the inter-layer convolutional features, while the self-interaction modules (SIMs) are proposed to produce multi-scale representation from the intra-layer features. The details of the SIMs can be seen in Fig.~\ref{fig:sim}. Similarly, we also apply the \textbf{transformation-interaction-fusion} strategy in the SIMs.
Concretely speaking, the resolution and dimension of the input feature are reduced by a convolutional layer, at first. In each branch, the SIM performs an initial \textbf{transformation} to adapt to the following interaction operation: We up-sample low-resolution features and sub-sample high-resolution features to the same resolution as the features from the other branch. The \textbf{interaction} between high- and low-resolution features with different channel numbers can obtain plenty of knowledge about various scales and maintain high-resolution information with a low parameter quantity. For ease of optimization, a residual connection is also adopted as shown in Fig.~\ref{fig:sim}. After up-sampling, normalization, and nonlinear processing, an FU is used to \textbf{fuse} the features of double paths from the SIM and the residual branch.
Integrating the SIMs into the decoder allows the network to adaptively deal with scale variation of different samples during the training stage. The entire process is written as:
\begin{equation}
    f_{SIM}^i = f_{add}^i + \textbf{M}^i_{SIM} \left ( \textbf{B}_{SIM}^{i,0}(f_{add}^{i}) + \textbf{B}_{SIM}^{i,1}(f_{add}^{i}) \right ),
    \label{equ:sim}
\end{equation}

\noindent where $f_{SIM}^i$ is the output of the $\mathbf{SIM}^i$. $\mathbf{M}(\cdot)$ represents the branch merging and $\textbf{B}_{SIM}^{i,j}$ denotes the operation in the $j^{th}$ branch (i.e.\ $\textit{B}^j$) in the $\mathbf{SIM}^i$, and the input feature $f_{add}^i$ is calculated as follows:
\begin{equation}
    \begin{split}
        f_{add}^i = \left \{ \begin{array}{ll}
            f_{AIM}^i + \textbf{U}^{i+1}(\textbf{F}^{i+1}(f_{SIM}^{i+1})) & \text{ if } i=0, 1, 2, 3, \\
            f_{AIM}^i                                                     & \text{ if } i = 4,
        \end{array} \right.
    \end{split}
    \label{equ:fadd}
\end{equation}

\noindent where $\textbf{U}^{i+1}(\cdot)$ and $\textbf{F}^{i+1}(\cdot)$ denote the $(i+1)^{th}$ up-sampling operation and fusion unit in the top-down pathway. For more details about the SIMs, please see Fig.~\ref{fig:sim}.

\subsection{Consistency-Enhanced Loss}\label{sec:cel}

In the SOD task, the widely used binary cross entropy function accumulates the per-pixel loss in the whole batch and does not consider the inter-pixel relationships, which can not explicitly prompt the model to highlight the foreground region as smoothly as possible and deal well with the sample imbalance issue. To this end, we propose a consistency-enhanced loss (CEL).
First of all, the final prediction is calculated as follows:
\begin{equation}
    \mathcal{P} = \text{Sigmoid}(\text{Conv}(\textbf{F}^0(f_{add}^0))),
    \label{equ:prediction}
\end{equation}

\noindent where $\mathcal{P} \in \mathbb{R}^{N \times H \times W \times 1}$ denotes $N$ saliency maps in a batch, and $N$ is the batchsize. $0 < {p} \in \mathcal{P} < 1$ is the probability of belonging to salient regions.
$\text{Sigmoid}(\text{Conv}(\cdot))$ actually represents the last convolutional layer with a nonlinear activation function in the decoder. The binary cross entropy loss (BCEL) function is written as follows:
\begin{equation}
    L_{BCEL} = \sum_{p\in \mathcal{P}, g\in \ \mathcal{G} } -\left [ g \log p + (1 - g) \log (1 - p) \right ],
    \label{equ:bcel}
\end{equation}

\noindent where $\log(\cdot)$ is also an element-wise operation. $\mathcal{G} \in \{0, 1\}^{N \times H \times W \times 1}$ represents the ground truth.
To address the fore-/back-ground imbalance issue caused by various scales, the loss function needs to meet two requirements, at least:
1) It focuses more on the foreground than the background, and the difference at the scale of objects does not induce the wide fluctuation in the computed loss; 2) When the predicted foreground region is completely disjoint from the ground-truth one, there should be the largest penalty. Based on the two points, we consider the topological relationships among regions to define the CEL as follows:
\begin{equation}
    \begin{split}
        L_{CEL} & = \frac{|FP + FN|}{|FP + 2TP + FN|} \\
        & = \frac{\sum (p-pg) + \sum (g-pg)}{\sum p + \sum g},
    \end{split}
    \label{equ:cel}
\end{equation}

\noindent where $TP$, $FP$ and $FN$ represent true-positive, false-positive and false-negative, respectively. $|\cdot|$ computes the area. $FP + FN$ denotes the difference set between the union and intersection of the predicted foreground region and the ground-truth one, while $FP + 2TP + FN$ represents the sum of this union set and this intersection. When $\{p | p > 0, p \in \mathcal{P}\}  \cap  \{g | g = 1, g \in \mathcal{G}\} = \varnothing$, the loss reaches its maximum, i.e.\ $L_{CEL} = 1$. Since $p$ is continuous, $L_{CEL}$ is differentiable with reference to $p$. Thus, the network can be trained in an end-to-end manner.

To compare $L_{CEL}$ with $L_{BCEL}$, we analyze their gradients which directly act on the network predictions. Their derivatives are expressed as follows:
\begin{equation}
    \frac{\partial L_{BCEL}}{\partial p} = -\frac{g}{p} + \frac{1 - g}{1 - p},
    \label{dbcel}
\end{equation}
\begin{equation}
    \begin{split}
        \frac{\partial L_{CEL}}{\partial p} = \frac{1-2g}{\sum (p+g)} - \frac{\sum (p+g-2pg)}{[\sum (p+g)]^2}.
    \end{split}
    \label{dcel}
\end{equation}

It can be observed that $\partial L_{BCEL} / \partial p$ only relies on the prediction of the individual position. While $\partial L_{CEL} / \partial p$ is related to all pixels in both the prediction $\mathcal{P}$ and the ground truth $\mathcal{G}$. Therefore, the CEL is considered to enforce a global constraint on the prediction results, which can produce more effective gradient propagation. In Equ. (\ref{dcel}), except that the numerator term $1-2g$ is position-specific, the other terms are image-specific. And this numerator is closely related to the binary ground truth, which results in that the inter-class derivatives have large differences while the intra-class ones are relatively consistent. This has several merits: 1) It ensures that there is enough large gradient to drive the network in the later stage of training; 2) It helps solve the intra-class inconsistency and inter-class indistinction issues, to some extent, thereby promoting the predicted boundaries of salient objects to become sharper.
Finally, the total loss function can be written as:
\begin{equation}
    L = L_{BCEL}(\mathcal{P}, \mathcal{G}) + \lambda L_{CEL} (\mathcal{P}, \mathcal{G}),
    \label{equ:finalloss}
\end{equation}

\noindent where $\lambda$ is a hyperparameter that balances the contributions of the two losses. For the sake of simplicity, it is set to $1$.

\section{Experiments}

\begin{table*}[t]
    \centering
    \renewcommand\arraystretch{1.3}
    \caption{Quantitative evaluation. The maximum, mean and weighted F-measure (larger is better), E-measure (larger is better), S-measure (larger is better) and MAE (smaller is better) of different saliency detection methods on five benchmark datasets. The best three results are highlighted in {\color[HTML]{FE0000} \textbf{red}}, {\color[HTML]{34FF34} \textbf{green}} and {\color[HTML]{3166FF} \textbf{blue}}. $^{\ast}$: with post-processing; $^{X101}$: ResNeXt-101~\cite{ResNext} as backbone; $^{R101}$: ResNet-101~\cite{Resnet} as backbone.}
    \label{tab:totalresult}
    \resizebox{\textwidth}{!}{%
        \begin{tabular}{|l|cccccc|cccccc|cccccc|cccccc|cccccc|}
            \hline                   & \multicolumn{6}{c|}{DUTS-TE} & \multicolumn{6}{c|}{DUT-OMRON} & \multicolumn{6}{c|}{HKU-IS} & \multicolumn{6}{c|}{ECSSD} & \multicolumn{6}{c|}{Pascal-S}                                                                                                                                                                                                                                                                                                                                                                                                                                                                                                                                                                                                                         \\ \cline{2-31}
            \multirow{-2}{*}{Model}  & F$_{max}$                    & F$_{avg}$                      & F$^{\omega}_{\beta}$        & E$_{m}$                    & S$_m$                         & MAE                   & F$_{max}$             & F$_{avg}$             & F$^{\omega}_{\beta}$  & E$_{m}$               & S$_m$                 & MAE                   & F$_{max}$             & F$_{avg}$             & F$^{\omega}_{\beta}$  & E$_{m}$               & S$_m$                 & MAE                   & F$_{max}$             & F$_{avg}$             & F$^{\omega}_{\beta}$  & E$_{m}$               & S$_m$                 & MAE                   & F$_{max}$             & F$_{avg}$             & F$^{\omega}_{\beta}$  & E$_{m}$               & S$_m$                 & MAE                   \\ \hline
            \multicolumn{31}{|c|}{VGG-16}                                                                                                                                                                                                                                                                                                                                                                                                                                                                                                                                                                                                                                                                                                                                                                               \\ \hline
            Ours                    & \textBC{red}{0.877}          & \textBC{red}{0.823}            & \textBC{red}{0.813}         & \textBC{red}{0.912}        & \textBC{green}{0.875}         & \textBC{red}{0.039}   & \textBC{blue}{0.794}                 & \textBC{blue}{0.741}                 & \textBC{blue}{0.719}                 & \textBC{green}{0.864} & 0.822                 & \textBC{red}{0.057} & \textBC{red}{0.932}   & \textBC{red}{0.906}   & \textBC{red}{0.892}   & \textBC{red}{0.955}   & \textBC{red}{0.914}   & \textBC{red}{0.030}   & \textBC{red}{0.943}   & \textBC{red}{0.922}   & \textBC{red}{0.905}   & \textBC{red}{0.947}   & \textBC{red}{0.919}   & \textBC{red}{0.036}   & \textBC{red}{0.882}   & \textBC{red}{0.843}   & \textBC{red}{0.820}   & \textBC{red}{0.898}   & \textBC{red}{0.855}   & \textBC{red}{0.065}   \\ \hline
            EGNet$_{19}$             & \textBC{red}{0.877}          & \textBC{blue}{0.800}                          & \textBC{blue}{0.797}        & \textBC{blue}{0.895}                      & \textBC{red}{0.878}           & \textBC{blue}{0.044}  & \textBC{red}{0.809}   & \textBC{green}{0.744}  & \textBC{red}{0.728} & \textBC{green}{0.864} & \textBC{red}{0.836}   & \textBC{red}{0.057} & \textBC{green}{0.927}  & 0.893                 & \textBC{blue}{0.875}  & \textBC{blue}{0.950}  & \textBC{green}{0.910} & \textBC{blue}{0.035}  & \textBC{red}{0.943}   & \textBC{blue}{0.913}  & \textBC{blue}{0.892}                 & 0.941                 & \textBC{red}{0.919}   & \textBC{blue}{0.041}  & \textBC{blue}{0.871}                 & 0.821                 & 0.798                 & 0.873                 & 0.848                 & 0.078                 \\ \hline
            AFNet$_{19}$             & \textBC{blue}{0.863}                        & 0.793                          & 0.785                       & \textBC{blue}{0.895}                      & \textBC{blue}{0.867}                         & 0.046                 & \textBC{green}{0.797} & 0.739                 & 0.717                 & \textBC{blue}{0.860}  & \textBC{green}{0.826}  & \textBC{red}{0.057} & \textBC{blue}{0.925}                 & 0.889                 & 0.872                 & 0.949                 & 0.906                 & 0.036                 & \textBC{blue}{0.935}                 & 0.908                 & 0.886                 & \textBC{blue}{0.942}  & \textBC{green}{0.914} & 0.042                 & \textBC{blue}{0.871}                 & \textBC{blue}{0.828}                 & \textBC{blue}{0.804}                 & \textBC{green}{0.887} & \textBC{blue}{0.850}  & \textBC{green}{0.071} \\ \hline
            MLMSNet$_{19}$           & 0.852                        & 0.745                          & 0.761                       & 0.863                      & 0.862                         & 0.049                 & 0.774                 & 0.692                 & 0.681                 & 0.839                 & 0.809                 & 0.064                 & 0.920                 & 0.871                 & 0.860                 & 0.938                 & \textBC{blue}{0.}907                 & 0.039                 & 0.928                 & 0.868                 & 0.871                 & 0.916                 & 0.911                 & 0.045                 & 0.864                 & 0.771                 & 0.785                 & 0.847                 & 0.845                 & 0.075                 \\ \hline
            PAGE$_{19}$              & 0.838                        & 0.777                          & 0.769                       & 0.886                      & 0.854                         & 0.052                 & 0.792                 & 0.736                 & \textBC{green}{0.722}  & \textBC{blue}{0.860}  & \textBC{blue}{0.825}                 & \textBC{green}{0.062}  & 0.920                 & 0.884                 & 0.868                 & 0.948                 & 0.904                 & 0.036                 & 0.931                 & 0.906                 & 0.886                 & \textBC{green}{0.943} & \textBC{blue}{0.912}                 & 0.042                 & 0.859                 & 0.817                 & 0.792                 & 0.879                 & 0.840                 & 0.078                 \\ \hline
            HRS$_{19}$               & 0.843                        & 0.793                          & 0.746                       & 0.889                      & 0.829                         & 0.051                 & 0.762                 & 0.708                 & 0.645                 & 0.842                 & 0.772                 & 0.066                 & 0.913                 & 0.892                 & 0.854                 & 0.938                 & 0.883                 & 0.042                 & 0.920                 & 0.902                 & 0.859                 & 0.923                 & 0.883                 & 0.054                 & 0.852                 & 0.809                 & 0.748                 & 0.850                 & 0.801                 & 0.090                 \\ \hline
            CPD$_{19}$               & \textBC{green}{0.864}         & \textBC{green}{0.813}          & \textBC{green}{0.801}       & \textBC{green}{0.908}      & \textBC{blue}{0.867}                         & \textBC{green}{0.043} & \textBC{blue}{0.794}                 & \textBC{red}{0.745} & 0.715                 & \textBC{red}{0.868}   & 0.818                 & \textBC{red}{0.057} & 0.924                 & \textBC{blue}{0.896}                 & \textBC{green}{0.881} & \textBC{green}{0.952} & 0.904                 & \textBC{green}{0.033} & \textBC{green}{0.936}  & \textBC{green}{0.914} & \textBC{green}{0.894} & \textBC{green}{0.943} & 0.910                 & \textBC{green}{0.040} & \textBC{green}{0.873} & \textBC{green}{0.832} & \textBC{green}{0.806} & \textBC{blue}{0.884}  & 0.843                 & \textBC{blue}{0.074}                 \\ \hline
            C2SNet$_{18}$            & 0.811                        & 0.717                          & 0.717                       & 0.847                      & 0.831                         & 0.062                 & 0.759                 & 0.682                 & 0.663                 & 0.828                 & 0.799                 & 0.072                 & 0.898                 & 0.851                 & 0.834                 & 0.928                 & 0.886                 & 0.047                 & 0.911                 & 0.865                 & 0.854                 & 0.915                 & 0.896                 & 0.053                 & 0.857                 & 0.775                 & 0.777                 & 0.850                 & 0.840                 & 0.080                 \\ \hline
            RAS$_{18}$               & 0.831                        & 0.751                          & 0.740                       & 0.864                      & 0.839                         & 0.059                 & 0.787                 & 0.713                 & 0.695                 & 0.849                 & 0.814                 & \textBC{green}{0.062}  & 0.913                 & 0.871                 & 0.843                 & 0.931                 & 0.887                 & 0.045                 & 0.921                 & 0.889                 & 0.857                 & 0.922                 & 0.893                 & 0.056                 & 0.838                 & 0.787                 & 0.738                 & 0.837                 & 0.795                 & 0.104                 \\ \hline
            PAGR$_{18}$              & 0.854                        & 0.784                          & 0.724                       & 0.883                      & 0.838                         & 0.055                 & 0.771                 & 0.711                 & 0.622                 & 0.843                 & 0.775                 & 0.071                 & 0.919                 & 0.887                 & 0.823                 & 0.941                 & 0.889                 & 0.047                 & 0.927                 & 0.894                 & 0.834                 & 0.917                 & 0.889                 & 0.061                 & 0.858                 & 0.808                 & 0.738                 & 0.854                 & 0.817                 & 0.093                 \\ \hline
            PiCANet$_{18}$           & 0.851                        & 0.749                          & 0.747                       & 0.865                      & 0.861                         & 0.054                 & \textBC{blue}{0.794}                 & 0.710                 & 0.691                 & 0.842                 & \textBC{green}{0.826}  & 0.068                 & 0.922                 & 0.870                 & 0.848                 & 0.938                 & 0.905                 & 0.042                 & 0.931                 & 0.885                 & 0.865                 & 0.926                 & \textBC{green}{0.914} & 0.046                 & \textBC{blue}{0.871}                 & 0.804                 & 0.781                 & 0.862                 & \textBC{green}{0.851} & 0.077                 \\ \hline
            DSS$_{17}^{\ast}$        & 0.825                        & 0.789                          & 0.755                       & 0.885                      & 0.824                         & 0.056                 & 0.781                 & 0.740                 & 0.697                 & 0.844                 & 0.790                 & \textBC{blue}{0.063}                 & 0.916                 & \textBC{green}{0.902} & 0.867                 & 0.935                 & 0.878                 & 0.040                 & 0.899                 & 0.863                 & 0.822                 & 0.907                 & 0.873                 & 0.068                 & 0.843                 & 0.812                 & 0.762                 & 0.848                 & 0.795                 & 0.096                 \\ \hline
            UCF$_{17}$               & 0.773                        & 0.631                          & 0.596                       & 0.770                      & 0.782                         & 0.112                 & 0.730                 & 0.621                 & 0.574                 & 0.768                 & 0.760                 & 0.120                 & 0.888                 & 0.823                 & 0.780                 & 0.904                 & 0.874                 & 0.061                 & 0.903                 & 0.844                 & 0.806                 & 0.896                 & 0.884                 & 0.069                 & 0.825                 & 0.738                 & 0.700                 & 0.809                 & 0.807                 & 0.115                 \\ \hline
            MSRNet$_{17}$            & 0.829                        & 0.723                          & 0.720                       & 0.848                      & 0.839                         & 0.061                 & 0.782                 & 0.687                 & 0.670                 & 0.827                 & 0.808                 & 0.073                 & 0.914                 & 0.866                 & 0.853                 & 0.940                 & 0.903                 & 0.040                 & 0.911                 & 0.868                 & 0.850                 & 0.918                 & 0.895                 & 0.054                 & 0.858                 & 0.790                 & 0.769                 & 0.854                 & 0.841                 & 0.081                 \\ \hline
            NLDF$_{17}$              & 0.812                        & 0.739                          & 0.710                       & 0.855                      & 0.816                         & 0.065                 & 0.753                 & 0.684                 & 0.634                 & 0.817                 & 0.770                 & 0.080                 & 0.902                 & 0.872                 & 0.839                 & 0.929                 & 0.878                 & 0.048                 & 0.905                 & 0.878                 & 0.839                 & 0.912                 & 0.875                 & 0.063                 & 0.833                 & 0.782                 & 0.742                 & 0.842                 & 0.804                 & 0.099                 \\ \hline
            AMU$_{17}$               & 0.778                        & 0.678                          & 0.658                       & 0.803                      & 0.804                         & 0.085                 & 0.743                 & 0.647                 & 0.626                 & 0.784                 & 0.781                 & 0.098                 & 0.899                 & 0.843                 & 0.819                 & 0.915                 & 0.886                 & 0.050                 & 0.915                 & 0.868                 & 0.840                 & 0.912                 & 0.894                 & 0.059                 & 0.841                 & 0.771                 & 0.741                 & 0.831                 & 0.821                 & 0.098                 \\ \hline
            \multicolumn{31}{|c|}{ResNet-50/ResNet-101/ResNeXt-101}                                                                                                                                                                                                                                                                                                                                                                                                                                                                                                                                                                                                                                                                                                                                                     \\ \hline
            Ours-R                  & \textBC{blue}{0.884}         & \textBC{red}{0.828}            & \textBC{red}{0.825}         & \textBC{red}{0.917}        & \textBC{blue}{0.884}          & \textBC{red}{0.037}   & 0.810                 & \textBC{red}{0.756}   & \textBC{green}{0.738}                 & \textBC{green}{0.873}                 & 0.833                 & \textBC{green}{0.055}                 & \textBC{red}{0.935}   & \textBC{red}{0.908}   & \textBC{red}{0.899}   & \textBC{red}{0.961}   & \textBC{red}{0.920}   & \textBC{red}{0.028}   & \textBC{green}{0.947} & \textBC{red}{0.924}   & \textBC{red}{0.911}   & \textBC{red}{0.953}   & \textBC{green}{0.925} & \textBC{red}{0.033}   & \textBC{green}{0.882} & \textBC{red}{0.842}   & \textBC{red}{0.821}   & \textBC{red}{0.899}   & \textBC{green}{0.857} & \textBC{red}{0.064}   \\ \hline
            SCRN$_{19}$              & \textBC{green}{0.888}        & \textBC{blue}{0.809}           & 0.803                       & 0.901                      & \textBC{green}{0.885}         & \textBC{blue}{0.040}  & \textBC{blue}{0.811}  & 0.746                 & 0.720                 & \textBC{blue}{0.869}  & \textBC{green}{0.837} & \textBC{blue}{0.056}  & \textBC{red}{0.935}   & 0.897                 & 0.878                 & 0.954                 & \textBC{blue}{0.917}  & 0.033                 & \textBC{red}{0.950}   & 0.918                 & 0.899                 & 0.942                 & \textBC{red}{0.927}   & \textBC{blue}{0.037}  & \textBC{red}{0.890}   & \textBC{green}{0.839} & \textBC{blue}{0.816}  & \textBC{blue}{0.888}  & \textBC{red}{0.867}   & \textBC{green}{0.065} \\ \hline
            EGNet-R$_{19}$           & \textBC{red}{0.889}          & \textBC{green}{0.815}          & \textBC{green}{0.816}       & \textBC{green}{0.907}      & \textBC{red}{0.887}           & \textBC{green}{0.039}                 & \textBC{red}{0.815}                 & \textBC{red}{0.756}                 & \textBC{green}{0.738}                 & \textBC{red}{0.874}                 & \textBC{red}{0.841}                 & \textBC{red}{0.053}                 & \textBC{red}{0.935}                 & \textBC{green}{0.901}                 & \textBC{blue}{0.887}                 & \textBC{green}{0.956}                 & \textBC{green}{0.918}                 & \textBC{green}{0.031}                 & \textBC{green}{0.947}                 & \textBC{blue}{0.920}                 & 0.903                 & \textBC{blue}{0.947}                 & \textBC{green}{0.925}                 & \textBC{blue}{0.037}                 & 0.878                 & 0.831                 & 0.807                 & 0.879                 & 0.853                 & 0.075                 \\ \hline
            CPD-R$_{19}$             & 0.865                        & 0.805                          & 0.795                       & \textBC{blue}{0.904}       & 0.869                         & 0.043                 & 0.797                 & \textBC{blue}{0.747}  & 0.719                 & \textBC{green}{0.873} & 0.825                 & \textBC{blue}{0.056}  & \textBC{blue}{0.925}  & 0.891                 & 0.876                 & 0.952                 & 0.906                 & 0.034                 & 0.939                 & 0.917                 & 0.898                 & \textBC{green}{0.949} & 0.918                 & \textBC{blue}{0.037}  & 0.872                 & 0.831                 & 0.803                 & 0.887                 & 0.847                 & 0.072                 \\ \hline
            ICNet$_{19}$             & 0.855                        & 0.767                          & 0.762                       & 0.880                      & 0.865                         & 0.048                 & \textBC{green}{0.813} & 0.739                 & 0.730                 & 0.859                 & \textBC{green}{0.837} & 0.061                 & \textBC{blue}{0.925}  & 0.880                 & 0.858                 & 0.943                 & 0.908                 & 0.037                 & 0.938                 & 0.880                 & 0.881                 & 0.923                 & 0.918                 & 0.041                 & 0.866                 & 0.786                 & 0.790                 & 0.860                 & 0.850                 & 0.071                 \\ \hline
            BANet$_{19}$             & 0.872                        & \textBC{green}{0.815}          & \textBC{blue}{0.811}        & \textBC{green}{0.907}      & 0.879                         & \textBC{blue}{0.040}  & 0.803                 & 0.746                 & \textBC{blue}{0.736}  & 0.865                 & 0.832                 & 0.059                 & \textBC{green}{0.930} & \textBC{blue}{0.899}  & \textBC{blue}{0.887}  & \textBC{blue}{0.955}  & 0.913                 & \textBC{blue}{0.032}  & \textBC{blue}{0.945}  & \textBC{green}{0.923} & \textBC{green}{0.908} & \textBC{red}{0.953}   & \textBC{blue}{0.924}  & \textBC{green}{0.035} & \textBC{blue}{0.879}  & \textBC{blue}{0.838}  & \textBC{green}{0.817} & \textBC{green}{0.889} & 0.853                 & \textBC{blue}{0.070}  \\ \hline
            BASNet$_{19}$            & 0.859                        & 0.791                          & 0.803                       & 0.884                      & 0.866                         & 0.048                 & 0.805                 & \textBC{red}{0.756}   & \textBC{red}{0.751}   & \textBC{blue}{0.869}  & \textBC{blue}{0.836}  & \textBC{blue}{0.056}  & \textBC{green}{0.930} & 0.898                 & \textBC{green}{0.890} & 0.947                 & 0.908                 & 0.033                 & 0.942                 & 0.879                 & \textBC{blue}{0.904}  & 0.921                 & 0.916                 & \textBC{blue}{0.037}  & 0.863                 & 0.781                 & 0.800                 & 0.853                 & 0.837                 & 0.077                 \\ \hline
            CapSal$_{19}^{R101}$     & 0.823                        & 0.755                          & 0.691                       & 0.866                      & 0.815                         & 0.062                 & 0.639                 & 0.564                 & 0.484                 & 0.703                 & 0.674                 & 0.096                 & 0.884                 & 0.843                 & 0.782                 & 0.907                 & 0.850                 & 0.058                 & 0.862                 & 0.825                 & 0.771                 & 0.866                 & 0.826                 & 0.074                 & 0.869                 & 0.827                 & 0.791                 & 0.878                 & 0.837                 & 0.074                 \\ \hline
            R3Net$_{18}^{\ast X101}$ & 0.833                        & 0.787                          & 0.767                       & 0.879                      & 0.836                         & 0.057                 & 0.795                 & \textBC{green}{0.748} & 0.728                 & 0.859                 & 0.817                 & 0.062                 & 0.915                 & 0.894                 & 0.878                 & 0.945                 & 0.895                 & 0.035                 & 0.934                 & 0.914                 & 0.902                 & 0.940                 & 0.910                 & 0.040                 & 0.846                 & 0.805                 & 0.765                 & 0.846                 & 0.805                 & 0.094                 \\ \hline
            DGRL$_{18}$              & 0.828                        & 0.794                          & 0.774                       & 0.899                      & 0.842                         & 0.050                 & 0.779                 & 0.709                 & 0.697                 & 0.850                 & 0.810                 & 0.063                 & 0.914                 & 0.882                 & 0.865                 & 0.947                 & 0.896                 & 0.038                 & 0.925                 & 0.903                 & 0.883                 & 0.943                 & 0.906                 & 0.043                 & 0.860                 & 0.814                 & 0.792                 & 0.881                 & 0.839                 & 0.075                 \\ \hline
            PiCANet-R$_{18}$         & 0.860                        & 0.759                          & 0.755                       & 0.873                      & 0.869                         & 0.051                 & 0.803                 & 0.717                 & 0.695                 & 0.848                 & 0.832                 & 0.065                 & 0.919                 & 0.870                 & 0.842                 & 0.941                 & 0.905                 & 0.044                 & 0.935                 & 0.886                 & 0.867                 & 0.927                 & 0.917                 & 0.046                 & 0.870                 & 0.804                 & 0.782                 & 0.862                 & \textBC{blue}{0.854}  & 0.076                 \\ \hline
            SRM$_{17}$               & 0.826                        & 0.753                          & 0.722                       & 0.867                      & 0.836                         & 0.059                 & 0.769                 & 0.707                 & 0.658                 & 0.843                 & 0.798                 & 0.069                 & 0.906                 & 0.873                 & 0.835                 & 0.939                 & 0.887                 & 0.046                 & 0.917                 & 0.892                 & 0.853                 & 0.928                 & 0.895                 & 0.054                 & 0.850                 & 0.804                 & 0.762                 & 0.861                 & 0.833                 & 0.085                 \\ \hline
        \end{tabular}%
    }
\end{table*}

\subsection{Datasets}

We evaluate the proposed model on five benchmark datasets: DUTS~\cite{DUTS}, DUT-OMRON~\cite{DUT-OMRON}, ECSSD~\cite{ECSSD}, HKU-IS~\cite{HKU-IS}, and PASCAL-S~\cite{PASCAL-S}. The DUTS contains 10,553 training and 5,019 test images, which is currently the largest salient object detection dataset. Both training and test sets contain complicated scenes. The DUT-OMRON contains 5,168 images of complex backgrounds and high content variety. The ECSSD is composed of 1,000 images with structurally complex natural contents. The HKU-IS contains 4,447 complex scenes that contain multiple disconnected objects with relatively diverse spatial distributions, and a similar fore-/back-ground appearance makes it more difficult to distinguish.
We follow the data partition of~\cite{HKU-IS,DSS,RAS} to use 1,447 images for testing.
The PASCAL-S consists of 850 challenging images.

\subsection{Evaluation Criteria}

In this paper, we use six measurements to eveluate every models.
\textbf{Precision-Recall (PR) curve.} We binarize the gray-scale prediction by a fixed threshold. The resulted binary map and the ground truth are used to calculate $\text{Precision} = \text{TP} / (\text{TP} + \text{FP})$ and $\text{Recall} = \text{TP} / (\text{TP} + \text{FN})$, where TP, FP and FN represent true-positive, false-positive and false-negative, respectively. The PR curve can be plotted by a group of pairs of precision and recall generated when the threshold slides from 0 to 255. The larger the area under the PR curve, the better the performance.
\textbf{F-measure~\cite{Fmeasure}, F$_{\beta}$} is formulated as the weighted harmonic mean of Precision and Recall~\cite{Fmeasure}, e.t.\ F$_\beta = \frac{(1+\beta^2)\text{Precision} \times \text{Recall}}{\beta^2 \text{Precision} + \text{Recall}}$, where $\beta^2$ is generally set to 0.3 to emphasize more on the precision. We calculate the maximal F$_\beta$ values from the PR curve, denoted as F$_{max}$ and use an adaptive threshold that is twice the mean value of the prediction to calculate F$_{avg}$. And the F$_{avg}$ can reflect the spatial consistency of the predictions~\cite{CPD}.
\textbf{MAE~\cite{MAE}} directly evaluates the average pixel-level relative error between the normalized prediction and the ground truth by calculating the mean of the absolute value of the difference.
\textbf{S-measure~\cite{Smeasure}, S$_{m}$} computes the object-aware and region-aware structure similarities, denoted as S$_o$ and S$_r$, between the prediction and the ground truth. S-measure is written as follows: $\text{S}_m = \alpha \cdot \text{S}_o + (1-\alpha) \cdot \text{S}_r$, where $\alpha$ is set to 0.5~\cite{Smeasure}.
\textbf{E-measure~\cite{Emeasure}, E$_{m}$} combines local pixel values with the image-level mean value to jointly evaluate the similarity between the prediction and the ground truth.
\textbf{weighted F-measure~\cite{wFmeasure}, F$^{\omega}_{\beta}$} defines a weighted precision, which is a measure of exactness, and a weighted recall, which is a measure of completeness is proposed to improve the existing metric to improve the existing metric
F-measure.

\begin{figure*}[ht]
    \centering
    \includegraphics[width=\linewidth]{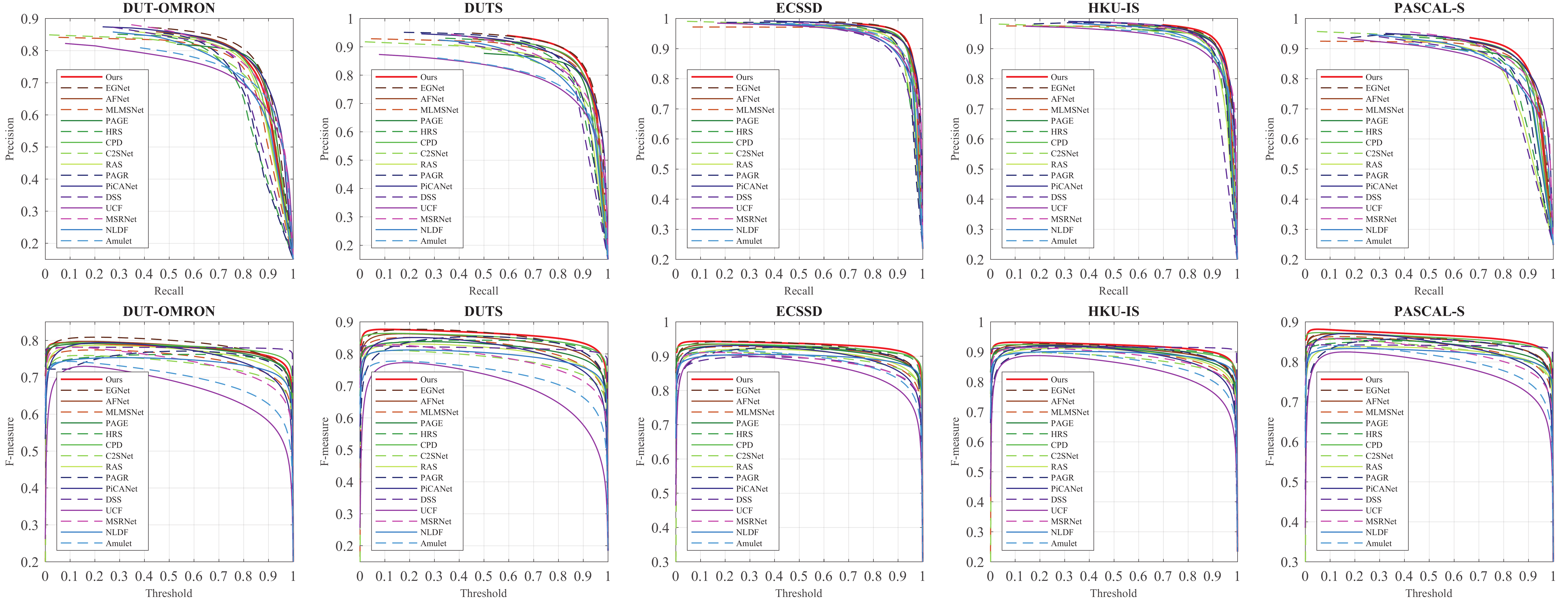}
    \caption{Precision-Recall curves ($1^{st}$ row) and F-measure curves ($2^{nd}$ row) on five common saliency datasets.}
    \label{fig:curves}
    \vspace{1em}
\end{figure*}
\begin{figure*}[]
    \centering
    \includegraphics[width=\linewidth]{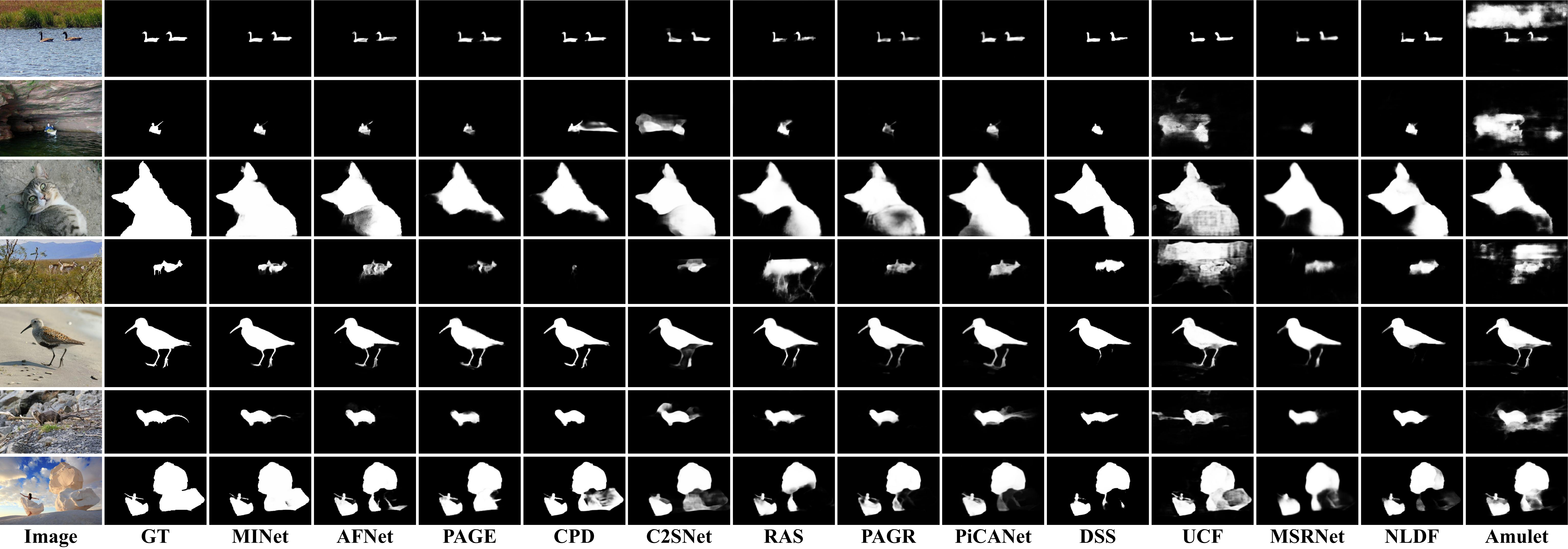}
    \caption{Visual comparisons of different methods.}
    \label{fig:comparesota}
\end{figure*}

\subsection{Implementation Details}

Following most existing state-of-the-art methods~\cite{BANet,BASNet,CPD,DGRL,PAGE-Net,HRS,BMPM,PiCANet,PAGRN,SRM,SCRN,EGNet}, we use the DUTS-TR~\cite{DUTS} as the training dataset. During the training stage, random horizontal flipping, random rotating, and random color jittering act as data augmentation techniques to avoid the over-fitting problem. To ensure model convergence, our network is trained for 50 epochs with a mini-batch of 4 on an NVIDIA GTX 1080 Ti GPU. The backbone parameters (i.e.\ VGG-16 and ResNet-50) are initialized with the corresponding models pretrained on the ImageNet dataset and the rest ones are initialized by the default setting of PyTorch. We use the momentum SGD optimizer with a weight decay of 5e-4, an initial learning rate of 1e-3 and a momentum of 0.9. Moreover, we apply a "poly" strategy~\cite{poly} with a factor of 0.9. The input size is $320 \times 320$.

\subsection{Comparison with State-of-the-arts}

We compare the proposed algorithm with 23 state-of-the-art saliency detection methods, including the SRM~\cite{SRM}, PiCANet~\cite{PiCANet}, DGRL~\cite{DGRL}, R3Net~\cite{R3Net}, CapSal~\cite{Capsal}, BASNet~\cite{BASNet}, BANet~\cite{BANet}, ICNet~\cite{ICNet}, CPD~\cite{CPD}, Amulet~\cite{Amulet}, NLDF~\cite{NLDF}, MSRNet~\cite{MSRNet}, UCF~\cite{UCF}, DSS~\cite{DSS}, PAGR\cite{PAGRN}, RAS~\cite{RAS}, C2SNet~\cite{C2SNet}, HRS~\cite{HRS}, PAGE~\cite{PAGE-Net}, MLMSNet~\cite{MLMSNet}, AFNet~\cite{AFNetRGB}, SCRN~\cite{SCRN}, and EGNet~\cite{EGNet}. For fair comparisons, all saliency maps of these methods are provided by authors or computed by their released codes.

\noindent\textbf{Quantitative Comparison.} To fully compare the proposed method with these existing models, the detailed experimental results in terms of six metrics are listed in Tab.~\ref{tab:totalresult}. As can be seen from the results, our approach has shown very good performance and significantly outperforms other competitors, although some methods~\cite{DSS,R3Net} use CRF~\cite{CRF} or other post-processing methods. The proposed method consistently performs better than all the competitors across all six metrics on most datasets. In particular, in terms of the MAE, the performance is averagely improved by 8.11\% and 7.30\% over the second-best method CPD~\cite{CPD} with the VGG-16 as the backbone and EGNet~\cite{EGNet} with the ResNet-50 as the backbone, respectively.
In addition, we demonstrate the standard PR curves and the F-measure curves in Fig.~\ref{fig:curves}. Our approach (red solid line) achieves the best results on the DUTS-TE, ECSSD, PASCAL-S and HKU-IS datasets and is also very competitive on the DUT-OMRON.

\noindent\textbf{Qualitative Evaluation.} Some representative examples are shown in Fig.~\ref{fig:comparesota}. These examples reflect various scenarios, including small objects ($1^{st}$ and $2^{nd}$ rows), low contrast between salient object and image background ($3^{rd}$ and $4^{th}$ rows), objects with threadlike parts ($5^{th}$ and $6^{th}$ rows) and complex scene ($6^{th}$ and $7^{th}$ rows). Moreover, these images contain small-/middle- and large-scale objects. It can be seen that the proposed method can consistently produce more accurate and complete saliency maps with sharp boundaries and coherent details.

\subsection{Ablation Study}\label{sec:ablationstudy}

\begin{table}[t]
    \centering
    \caption{Ablation analysis on the DUTS-TE dataset.}
    \label{tab:ablationstudy}
    \resizebox{\linewidth}{!}{%
        \begin{tabular}{|c|cccccc|}
            \hline
            Model          & F$_{max}$ & F$_{avg}$ & F$^{\omega}_{\beta}$ & E$_{m}$ & S$_m$ & MAE    \\ \hline
            Baseline       & 0.829     & 0.738     & 0.725                & 0.859   & 0.842 & 0.057  \\ \hline
            +AIMs          & 0.855     & 0.775     & 0.768                & 0.884   & 0.860 & 0.047  \\
            +Amulet-like   & 0.845     & 0.758     & 0.747                & 0.872   & 0.851 & 0.052  \\ \hline
            +SIMs          & 0.865     & 0.786     & 0.773                & 0.888   & 0.865 & 0.047  \\
            +PPM           & 0.847     & 0.762     & 0.753                & 0.875   & 0.856 & 0.050  \\
            +ASPP          & 0.859     & 0.777     & 0.767                & 0.880   & 0.861 & 0.048  \\ \hline
            +AIMs+SIMs     & 0.874     & 0.792     & 0.789                & 0.893   & 0.874 & 0.044  \\ \hline
            +AIMs+SIMs+CEL & 0.877     & 0.823     & 0.813                & 0.912   & 0.875 & 0.039  \\ \hline
        \end{tabular}%
    }
\end{table}

To illustrate the effectiveness of each proposed module, we conduct a detailed analysis next.

\begin{figure}[t]
    \centering
    \includegraphics[width=\linewidth]{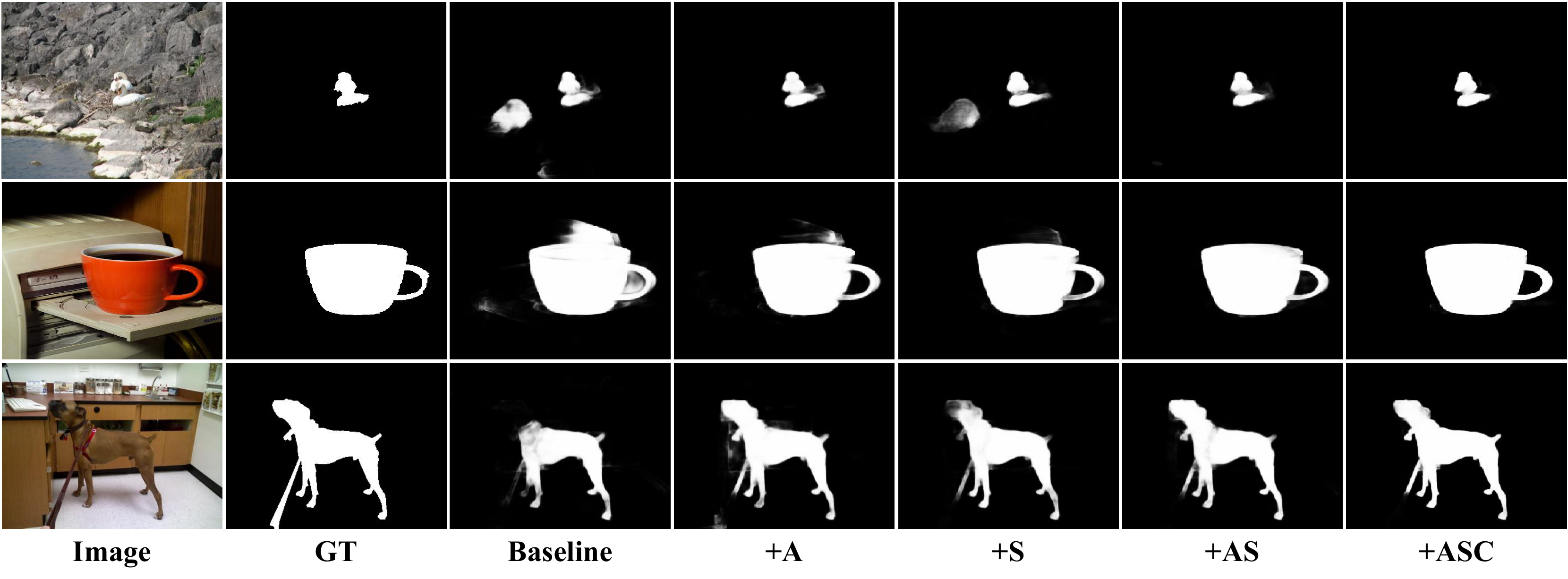}
    \caption{Visual comparisons for showing the benefits of the proposed modules. GT: Ground truth; A: AIMs; S: SIMs; C: CEL.}
    \label{fig:ablationstudy}
\end{figure}

\noindent\textbf{Effectiveness of the AIMs and SIMs.}
Our baseline model is an FPN-like network~\cite{FPN}, which uses the lateral connections to reduce the channel number to $32$ in the shallowest layer and to $64$ in the other layers. We separately install the AIMs and SIMs on the baseline network and evaluate their performance. The results are shown in Tab.~\ref{tab:ablationstudy}. It can be seen that both modules achieve significant performance improvement over the baseline. And, the proposed SIMs also performs much better than the PPM~\cite{PPM} and the ASPP~\cite{Deeplab} and it has increased by 6.21\% and 1.45\% in MAE, especially. In addition, the combination of the AIMs and SIMs can further improve the performance. The visual effects of different modules are illustrated in Fig.~\ref{fig:ablationstudy}. We can see that the AIMs and SIMs help effectively suppress the interference of backgrounds and completely segment salient objects because the richer multi-scale contextual information can be captured by the interactive feature learning.

\noindent\textbf{Comparisons with the Amulet-like~\cite{Amulet} strategy.} We compare the AIMs with the Amulet-like strategy in FLOPs, Parameters and GPU memory. ``+AIMs'': 137G, 47M and 1061MiB. ``+Amulet-like'': 176G, 20M and 1587MiB. AIMs combine fewer levels and have less computational cost. The fusion strategy achieves higher accuracy. And in Tab.~\ref{tab:ablationstudy}, it gets additional 2.14\%, 2.77\% and 8.26\% improvement in F$_{avg}$, F$^{\omega}_{\beta}$ and MAE over the model ``+Amulet-like''.

\noindent\textbf{Effectiveness of the CEL.}
We also quantitatively evaluate the effect of the consistency-enhanced loss (CEL) in Tab.~\ref{tab:ablationstudy}. Compared to ``+AIMs+SIMs'', the model with the CEL achieves consistent performance enhancements in terms of all six metrics. In particular, the F$_{avg}$, F$^{\omega}_{\beta}$ and MAE scores are respectively improved by 4.75\%, 3.75\%, and 13.16\%. Since the F$_{avg}$ is closely related to the spatial consistency of the predicted results~\cite{CPD}, the salient regions are more uniformly highlighted as shown in Fig.~\ref{fig:ablationstudy}.

\section{Conclusion}

In this paper, we investigate the multi-scale issue to propose an effective and efficient network MINet with the transformation-interaction-fusion strategy, for salient object detection. We first use the aggregate interaction modules (AIMs) to integrate the similar resolution features of adjacent levels in the encoder. Then, the self-interaction modules (SIMs) are utilized to extract the multi-scale information from a single level feature for the decoder. Both AIMs and SIMs interactively learn contextual knowledge from the branches of different resolutions to boost the representation capability of size-varying objects. Finally, we employ the consistency-enhanced loss (CEL) to alleviate the fore- and back-ground imbalance issue, which can also help uniformly highlight salient object regions. Each proposed module achieves significant performance improvement. Extensive experiments on five datasets validate that the proposed model outperforms 23 state-of-the-art methods under different evaluation metrics.

\section*{Acknowledgements}
This work was supported in part by the
National Key R\&D Program of China \#2018AAA0102003,
National Natural Science Foundation of China
\#61876202, \#61725202, \#61751212 and \#61829102,
and the Dalian Science and Technology Innovation Foundation \#2019J12GX039.

{\small
\bibliographystyle{ieee_fullname}
\bibliography{egbib}
}

\end{document}